\pdfoutput=1

\documentclass[11pt]{article}

\usepackage[final]{acl}

\usepackage{times}
\usepackage{latexsym}

\usepackage[T1]{fontenc}

\usepackage[utf8]{inputenc}

\usepackage{microtype}

\usepackage{inconsolata}

\usepackage{graphicx}
\usepackage{hyperref}
\usepackage{url}
\usepackage{graphicx}
\usepackage{enumitem}
\usepackage{bbding}
\usepackage{amsmath}
\usepackage{amssymb}
\usepackage{booktabs}
\usepackage{multirow}
\usepackage{wrapfig}
\usepackage{floatrow}
\newfloatcommand{capbtabbox}{table}[][\FBwidth]
\usepackage{xcolor}
\usepackage{listings}
\usepackage{color}
\usepackage{colortbl}
\usepackage{tcolorbox}
\usepackage[utf8]{inputenc}

\title{Grounded-VideoLLM: Sharpening Fine-grained Temporal Grounding in Video Large Language Models}

\author{Haibo Wang$^{1,6}$\thanks{Work done during an internship at UC Davis.}, \quad Zhiyang Xu$^2$,\quad  Yu Cheng$^3$,\quad  Shizhe Diao$^4$,\quad  Yufan Zhou$^5$,\\ 
\textbf{Yixin Cao$^6$},\quad \textbf{Qifan Wang$^7$},\quad  \textbf{Weifeng Ge$^6$},\quad \textbf{Lifu Huang$^1$} \thanks{Corresponding author. Codes will be available at \href{https://github.com/WHB139426/Grounded-Video-LLM}{https://github.com/WHB139426/Grounded-Video-LLM}} \\
$^1$University of California, Davis \quad $^2$Virginia Tech \quad $^3$The Chinese University of Hong Kong \\
$^4$NVIDIA \quad $^5$Adobe Research \quad $^6$Fudan University \quad $^7$Meta AI\\
\texttt{hibwang@ucdavis.edu, lfuhuang@ucdavis.edu} \\
}

\begin{document}


\maketitle

\begin{abstract}

Despite their impressive performance in coarse-grained video understanding, Video Large Language Models (Video-LLMs) still face challenges in fine-grained temporal grounding, including ineffective temporal modeling and inadequate timestamp representations. In this work, we introduce Grounded-VideoLLM, a novel Video-LLM designed to perceive and reason over specific video moments with fine-grained temporal precision. Our model features (1) a two-stream encoder that explicitly captures inter-frame relationships while preserving intra-frame visual details and (2) discrete temporal tokens enriched with structured time knowledge for timestamp representation. Besides, we propose a multi-stage training strategy tailored to such grounding-specific architecture. The model is initially trained on simple video-caption tasks and progressively introduced to complex video temporal grounding tasks, ensuring a smooth learning curve and temporal alignment. We further strengthen Grounded-VideoLLM’s temporal reasoning by constructing a VideoQA dataset with grounded information using an automated annotation pipeline. Extensive experiments demonstrate that Grounded-VideoLLM not only surpasses existing models in fine-grained grounding tasks but also exhibits strong potential as a general video understanding assistant.
\end{abstract}

\section{Introduction}

Multi-modal Large Language Models (MLLMs) have made remarkable progress in image-level understanding \citep{llava, instruct_blip, blip2}. However, extending their capabilities to the video domain poses distinct challenges. Unlike static images, the temporal nature of videos challenges models to process not only visual content but also the sequence and timing of events. While current Video-LLMs \citep{pllava, mvbench, video-llama, video-llava} are capable of capturing global visual semantics and generating coarse-grained captions for short clips, they struggle with fine-grained video understanding \citep{tempcompass, videohallucer}, which requires decomposing the video along the temporal axis to accurately perceive and reason over specific moments, such as subtle actions, transitions, and events that unfold over time.
    
\begin{figure*}
\label{fig:intro}
\begin{center}
\includegraphics[scale=0.47]{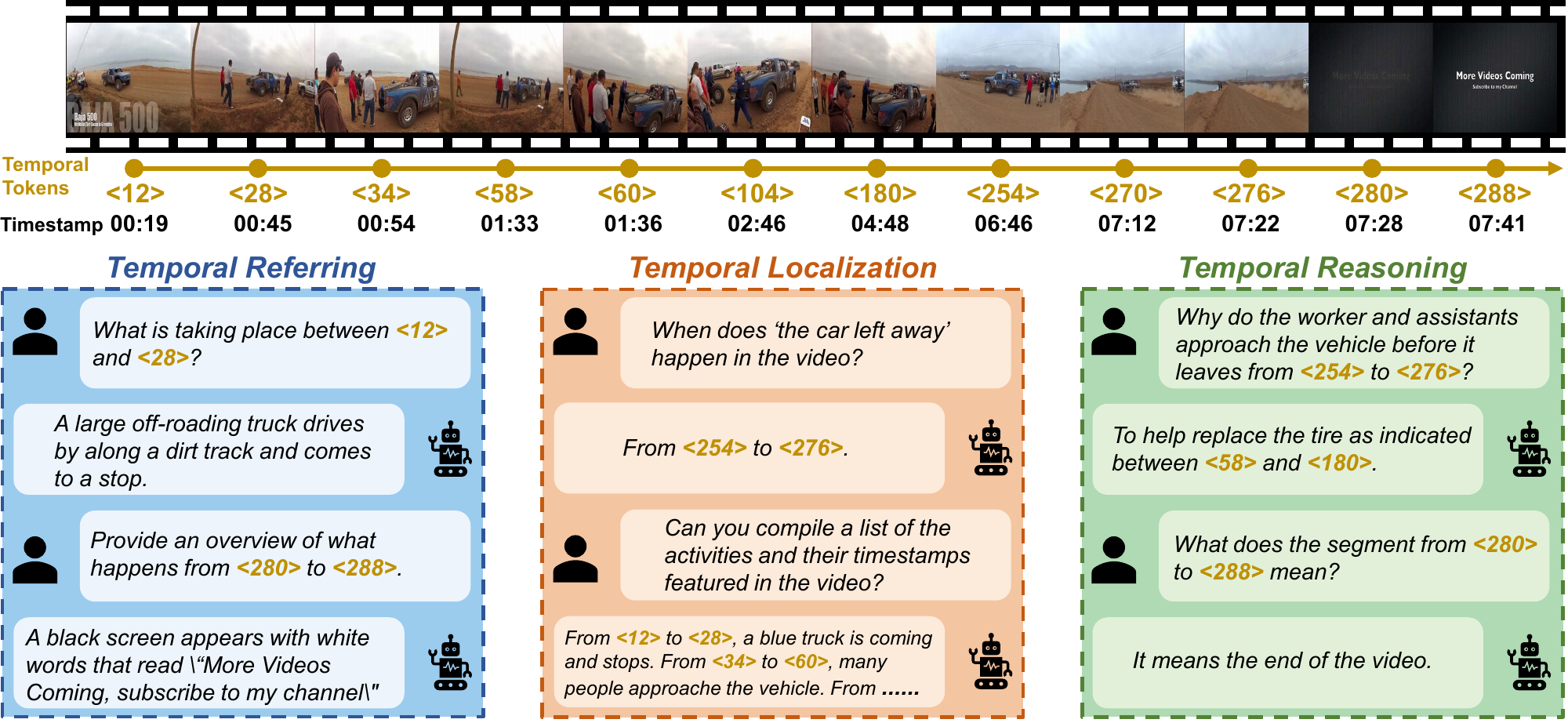}
\end{center}
\caption{\textit{Grounded-VideoLLM} enables Temporal Referring/Localizing/Reasoning for MLLMs.}
\label{fig:intro}
\end{figure*}

Previous research efforts \citep{timechat, vtimellm, momentor, lita, vtgllm} have explored \textit{temporal grounding} to improve fine-grained video understanding. However, two main challenges impede their potential for achieving effective \textit{fine-grained temporal grounding}: \textbf{(1)} Models like Video-ChatGPT \citep{video-chatgpt}, P-LLaVA \citep{pllava}, and Video-LLAMA \citep{video-llama} typically sample multiple frames from a video and encode each frame independently using an image encoder, followed by a feature projector (e.g., sliding Q-former \citep{timechat}, visual adapter \citep{vtimellm}). This focuses primarily on spatial details while potentially neglecting the temporal relationships between frames since their visual encoders are solely trained on images. \textbf{(2)} Current models also struggle with timestamp representation, which is crucial for pinpointing specific moments in time for fine-grained understanding. Models such as TimeChat \citep{timechat} and VTimeLLM \citep{vtimellm} represent timestamps as plain texts, for example, [{\ttfamily{"from 102.3 to 120.1 seconds"}}]. Despite being straightforward, this needs to tokenize continuous floating-point values, which is inefficient for LLMs since their next-token prediction paradigm struggles with handling numerical data \citep{numerologic, mathgpt}. Although there have been some previous works \citep{vid2seq,lita,momentor} using special tokens to represent time positions, Vid2Seq \cite{vid2seq} relies heavily on large-scale pre-training from scratch using noisy transcribed speech and is limited to the task of dense video captioning, while LITA \cite{lita} and Momentor \cite{momentor} only align these tokens with simple fine-tuning stage, which proves to be insufficient in our experiments (Table \ref{tab:ablation_two_stream}).

To further improve video understanding, we propose to sharpen the model with \textit{fine-grained temporal grounding}, allowing the model to recognize not only \textit{what} happens but pinpoint \textit{when} it happens with finer granularity, as illustrated in Figure \ref{fig:intro}. Targeting these goals, we introduce {\textit{Grounded-VideoLLM}}, a novel Video-LLM that can perceive and reason over specific video moments with fine-grained precision. \textbf{From the perspective of model architecture}, {\textit{Grounded-VideoLLM}} is built upon two key innovations: (1) Two-Stream Encoding: We decompose each segment of the video into spatial and temporal components and encode each with an expert encoder. The temporal stream extracts motion representations from dense frames and complements the spatial stream, which captures appearance representations. This dual-stream approach forms comprehensive video representations enriched with both temporal and spatial information. (2) Temporal Tokens: We extend the LLM’s vocabulary by introducing discrete tokens crafted to denote relative time positions and share a unified embedding space with the LLM, allowing \textit{Grounded-VideoLLM} to avoid the inefficiency of tokenizing numerical text and seamlessly predict both timestamps and textual outputs in a single sequence of discrete tokens. \textbf{From the perspective of training}, we start with an image MLLM \citep{phi3.5-vision} as the foundation and adopt a three-stage training strategy. We meticulously select different tasks for each stage, and progressively refine the model in a “coarse-to-fine” manner, transitioning from image understanding to video comprehension, and ultimately to fine-grained temporal grounding. This scheme ensures the introduced temporal tokens align closely with the video timelines and LLM's semantic space, distinguishing our method from previous studies \cite{lita, momentor, vid2seq}. Furthermore, we enhance the model's temporal reasoning by curating 17K grounded VideoQA \citep{nextgqa} samples with the assistance of GPT-4 \cite{gpt4}. Extensive experiments demonstrate that \textit{Grounded-VideoLLM} shows promising results over existing Video-LLMs not only in traditional video temporal grounding tasks but also in general video understanding benchmarks. 

\section{Related Work}
\textbf{Video Large Language Models} have caught a growing interest for general video understanding \citep{video-llama, video-llava}. However, they struggle with temporal perception \citep{tempcompass} and exhibit hallucination \citep{videohallucer} when asked about specific moments. They typically encode each frame independently using an image encoder to create video embeddings, resulting in video representations lacking inherent temporal information and heavily relying on the position embeddings of LLM for temporal understanding, limiting the model's capability to perform fine-grained temporal grounding. Norton \citep{lin2024Norton} employs video-paragraph and clip-caption contrastive losses to capture long-term dependencies for fine-grained alignment. In this work, we employ a two-stream architecture that integrates a video expert to extract motion features to complement the appearance features during the early encoding process. This is different from traditional two-stream networks \citep{two-stream, conv_two_stream} since we don't rely on heavy extraction of optical flows. Additionally, we employ a progressive training strategy that gradually adapts an image-based MLLM for fine-grained video understanding. Although concurrent works such as SlowFast-LLaVA \citep{slowfast-llava} and VideoGPT+ \citep{videogpt+} also introduce an additional stream, SlowFast-LLaVA relies on a single image encoder to process each video frame without training, missing crucial temporal relationships between frames. VideoGPT+ merely arranges video tokens as a prefix to image tokens using sparse frames. Instead, our approach is specifically designed for fine-grained temporal grounding, leveraging a unique encoding, pooling, and training strategy tailored for dense frames, along with a dedicated grounding mechanism.

\textbf{Video Temporal Grounding (VTG)} tasks usually include Temporal Sentence Grounding \citep{charades-sta, localizing}, Dense Video Captioning \citep{activitynet, youcook}, and Grounded VideoQA \citep{nextgqa}. Given the emerging capabilities of Video-LLMs \citep{video-llama, video-llava, internvideo2}, many studies have investigated how to adapt them for VTG tasks. For example, TimeChat \citep{timechat} and VTimeLLM \citep{vtimellm} perform temporal grounding using a fully text-to-text approach through instruction-tuning datasets. Momentor \citep{momentor} introduces a temporal perception module to address the quantization errors, and VTG-LLM \citep{vtgllm} incorporates absolute-time tokens to handle timestamps. Compared to these, we avoid textual representation of timestamps and instead introduce discrete temporal tokens for timestamp encoding. Different from previous methods that also use special tokens \cite{lita, momentor, vid2seq, kosmos2}, our model is more efficient by continuing training based on an established image MLLM with a two-stream architecture and a progressive training strategy.

\begin{figure*}
\label{fig:method}
\begin{center}
\includegraphics[scale=0.58]{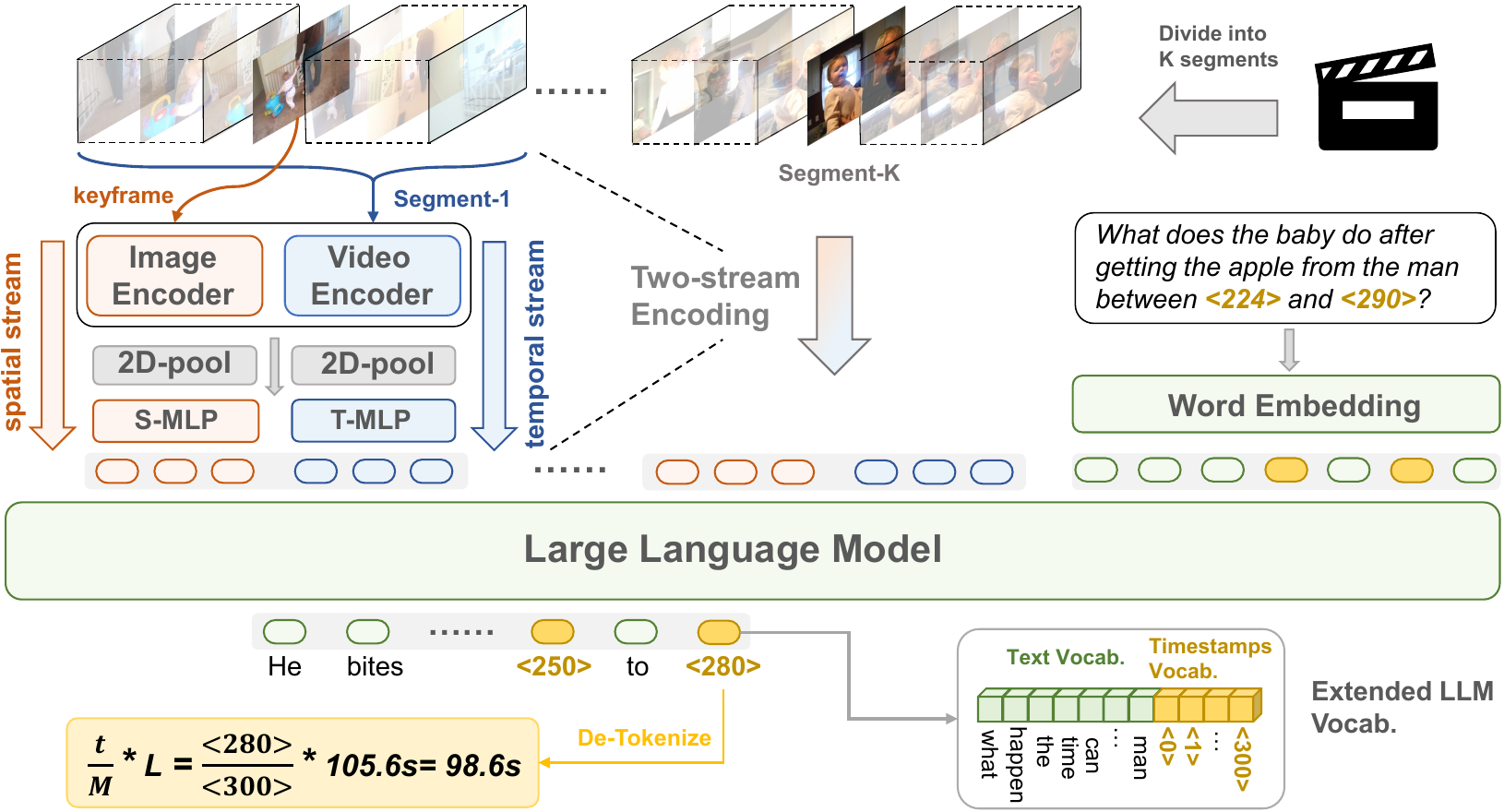}
\end{center}
\caption{Overview of \textit{Grounded-VideoLLM}. For temporal modeling, we employ a segment-wise encoding strategy by decomposing each segment into a spatial part and a temporal part and encoding each respectively. For timestamp representation, we introduce additional special temporal tokens sharing a unified embedding space with LLM.}
\label{fig:method}
\end{figure*}

\section{Model Architecture} 
Given that current MLLMs already exhibit strong image-understanding capabilities, our architecture aims to sharpen temporal awareness by capturing motion dynamics across frames, which serve as a vital supplement to spatial content. As shown in Figure \ref{fig:method}, we develop \textit{Grounded-VideoLLM} upon a well-established MLLM for spatial comprehension and integrate an expert video encoder for temporal comprehension. Additionally, to avoid tokenizing numerical texts, we incorporate temporal tokens into the LLM's vocabulary for efficient and unified timestamp representation.

\subsection{Two-Stream Encoding}
Given a video $\mathcal{V}$ with $T$ frames, we divide it into $K$ segments and employ a segment-wise encoding strategy. Due to the inherent redundancy of consecutive frames, each segment can be naturally represented from two perspectives: spatial and temporal. The spatial representation of each segment is derived from an individual keyframe, capturing the primary appearance semantics, while the temporal representation is learned from multiple frames depicting the motion evolution within the segment.


\textbf{Spatial Stream.} We sample the middle frame from each segment as the keyframe and extract its spatial features using the original image encoder from the MLLM \citep{clip}, resulting in spatial features $\textbf{F}_{S} \in \mathbb{R}^{H_S \times W_S \times D_S}$, where $H_S$, $W_S$, $D_S$ denote the height, width and dimension of the spatial features. Since dense frames are crucial for fine-grained temporal grounding, an appropriate pooling strategy is required to reduce token length. As indicated by \cite{pllava} and \cite{deco} that a 2D average pooling is both efficient and robust for spatial downsampling, we employ a 2D pooling kernel with a size $\sigma_{S} \times \sigma_{S}$ over the feature map and gets $\textbf{F}_{S} \in \mathbb{R}^{N_S \times D_S}$ as the feature for spatial stream, where $N_S = \frac{H_S}{\sigma_{S}} \times \frac{W_S}{\sigma_{S}}$.

\textbf{Temporal Stream.} Traditional two-stream networks typically encode the optical flow as the temporal stream. However, given the scale of data and parameters of MLLMs, extracting optical flow is computationally expensive and impractical. Consequently, we resort to a strong and well pre-trained video encoder to extract motion representations for each segment, using a lower resolution but more frames. We input each segment, containing $\frac{T}{K}$ frames, into the video encoder to obtain the segment-level features $\textbf{F}_{T} \in \mathbb{R}^{\frac{T}{K} \times H_T \times W_T \times D_T}$, where $H_T$, $W_T$, $D_T$ denote the height, width and dimension of each frame feature. Similar to the spatial stream, we apply a 2D average pooling strategy to downsample $\textbf{F}_{T}$. However, as the temporal stream focuses primarily on temporal modeling, we retain the complete temporal information by only pooling along the spatial dimensions. Specifically, we aggressively downsample $\textbf{F}_{T}$ using a kernel with a larger size of $\sigma_{T} \times \sigma_{T}$, resulting in the compressed $\textbf{F}_{T} \in \mathbb{R}^{\frac{T}{K} \times N_T \times D_T}$ for temporal stream, where $N_T = \frac{H_T}{\sigma_{T}} \times \frac{W_T}{\sigma_{T}}$. To get the complete segment-level representation, we flatten the features of both stream and concat them together :
\begin{equation}
\begin{aligned}
\textbf{F}_{Seg} = \operatorname{Concat} \left[f(\textbf{F}_{S}); g(\textbf{F}_{T}) \right]
\end{aligned}
\end{equation}
where $\textbf{F}_{Seg} \in \mathbb{R}^{(N_S + \frac{T}{K} \cdot N_T) \times D}$, $f(\cdot)$ and $g(\cdot)$ are two MLPs that project the visual features to LLM's dimension $D$. The final video representation is formed by concatenating the $K$ segment-level representations $\textbf{F}_{Seg}$, resulting in $\textbf{F}_{Vid} \in \mathbb{R}^{K \cdot (N_S + \frac{T}{K} \cdot N_T) \times D}$. This representation retains detailed spatial information across all segments along with their global temporal contexts, while maintaining a manageable token length. The combined video representation $\textbf{F}_{Vid}$ is then fed into the LLM serving as soft prompts, alongside the word embeddings of the instruction text $\textbf{F}_{Text}$ to generate the target response $\mathcal{A}$. The model is trained using the standard cross-entropy loss function for LLM with trainable parameters $\theta$: 
\begin{equation}
\begin{aligned}
  \mathcal{L} = -\sum_{t=1}^{L_{a}}logP_{\theta}(\mathcal{A}_t|\mathcal{A}_{<t}, \textbf{F}_{Vid}, \textbf{F}_{Text})
\label{eq:loss}
\end{aligned}
\end{equation}
where $\mathcal{A}_t$ is predicted autoregressively at position $t$, and $L_a$ is the sequence length of the ground truth answer text $\mathcal{A}$.

\subsection{Unified Temporal Tokens}
\label{subsec: temporal token}
Given a text depicting a video and its associated timestamps, we employ a relative time representation that converts continuous timestamps into a sequence of discrete tokens. For a video $\mathcal{V}$ with a duration of $L$ seconds, we evenly divide $\mathcal{V}$ into $M$ equal-length chunks, and then define $M+1$ anchor points (ranging from {\ttfamily{<0>}} to {\ttfamily{<M>}}) across $\mathcal{V}$, denoting relative temporal positions. Each anchor point corresponds to a specific timestamp and is encoded as a temporal token. For instance, {\ttfamily{<0>}} denotes the very start of $\mathcal{V}$ while {\ttfamily{<M>}} indicates the end. These $M+1$ tokens are added to the LLM's vocabulary to enable unified modeling alongside text. A specific continuous timestamp $\tau$ can be easily converted to a temporal token {\ttfamily{<t>}} and vice verse:
\begin{equation}
\begin{aligned}
t = \operatorname{Round}(M\cdot \frac{\tau}{L}), \ \ \ \tau = L\cdot \frac{t}{M}
\label{eq:timestamp}
\end{aligned}
\end{equation}
While this may introduce minor quantization errors, these can be minimized by selecting an appropriate $M$ or an interpolation expansion. We then organize the text span and its corresponding temporal tokens in a unified format. Both text tokens and temporal tokens are mapped to embeddings through the extended word embedding layer of LLM. For example, one input representation is as follows:
\begin{tcolorbox}
{\ttfamily{<video>}}$\textbf{F}_{Vid}${\ttfamily{</video>}} {\ttfamily{<grounded>}} From {\ttfamily{<0>}} to {\ttfamily{<6>}}, a baby is crying. From {\ttfamily{<7>}} to {\ttfamily{<16>}}, a man is coming and picking up the baby. From {\ttfamily{<20>}} to {\ttfamily{<25>}}, the baby is eating an apple.
\end{tcolorbox} 
where {\ttfamily{<video>}} and {\ttfamily{</video>}} represent the beginning and end of encoded video representations. {\ttfamily{<grounded>}} is a special token to tell the model should output the grounded timestamps.

\begin{table*}\scriptsize
\resizebox{\linewidth}{!}{
    \begin{tabular}{l|c|c|c}
    \toprule
    \textbf{Training Stage} & \textbf{Task} &\textbf{\# of Samples} & \textbf{Datasets}\\
    \midrule 
    Video-Caption Alignment & \multirow{1}{*}{Video Captioning} &1.28M &WebVid-10M, Panda-70M, InternVid-10M  \\
    \midrule 
    \multirow{3}{*}{Temporal Token Alignment} 
    & \cellcolor{gray!10} Temporal Sentence Grounding & \cellcolor{gray!10} 149K & \cellcolor{gray!10} VTimeLLM-Stage2  \\
    & \cellcolor{gray!10} Dense Video Captioning & \cellcolor{gray!10} 92K & \cellcolor{gray!10} VTimeLLM-Stage2, Moment-10M, InternVid-G \\
    & \cellcolor{gray!10} Temporal Referring & \cellcolor{gray!10} 95K & \cellcolor{gray!10} VTimeLLM-Stage2, InternVid-G \\
    \midrule 
    \multirow{9}{*}{Multi-Task Instruction Tuning} 
    & \cellcolor{gray!10} Grounded Conversation & \cellcolor{gray!10} 442K & \cellcolor{gray!10} RTL, Moment-10M  \\
    & \cellcolor{gray!10} Temporal Sentence Grounding  & \cellcolor{gray!10} 84K &\cellcolor{gray!10} DiDeMo, HiREST, QuerYD, VTG-IT \\
    & \cellcolor{gray!10} Dense Video Caption & \cellcolor{gray!10} 41K & \cellcolor{gray!10} COIN, ViTT, YouCook2, VTG-IT \\
    & \cellcolor{gray!10} Grounded VideoQA & \cellcolor{gray!10} 17K & \cellcolor{gray!10} Self Collected\\
    & Converstation &233K &VCG-Plus-112K, Videochatgpt-100K, Videochat2-Conv\\
    & VideoQA &282K &EgoQA, NExT-QA, Intent-QA, STAR, CLEVRER, WebVid-QA\\
    & Classification &66K &SthSthV2, Kinetics\\   
    & Video Captioning &136K &TextVR, YouCook2, WebVid, ShareGPT4Video\\   
                                          
    \bottomrule 
\end{tabular}
}
\caption{Datasets at three training stages. Tasks with \colorbox{gray!10}{gray} consist of datasets regarding temporal grounding.}
\label{tab:data}
\end{table*}

This strategy avoids the need to tokenize and process numerical values, which has been identified as a limitation of LLMs \citep{numerologic}. Notably, special tokens in LLMs are widely used in various domains. For example, Pix2Seq \citep{pix2seq} leverages special tokens to represent spatial grounding in images, Open-VLA \citep{kim2024openvla} and RT-2 \citep{brohan2023rt-2} utilize them for encoding robot action spaces, Yo'LLaVA \citep{nguyen2024yo-llava} use special tokens to refer to personalized subjects, and, most relevant to our work, Vid2Seq employs special tokens for temporal grounding in videos. However, the aforementioned works \textit{lack effective strategies to align these tokens with both LLM's semantic meanings and specific functionalities}. For example, Vid2Seq \citep{vid2seq} simply appends temporal tokens as a prefix to the caption and trains the entire T5 model from scratch using noised transcribed speech, which disrupts the language model’s original semantic embedding. In contrast, we introduce a temporal token alignment training stage in Sec. \ref{sec: training} to mitigate this issue. Instead of training the entire model, we update only the word embeddings of temporal tokens and the final logit head, with carefully curated grounding-specific datasets. This ensures that temporal tokens are aligned with both the video timeline and the LLM’s semantic space, and timestamps and text can be jointly decoded as a single sequence while maintaining the general video understanding ability.

\begin{figure*}[h]
\label{fig:data}
\begin{center}
\includegraphics[scale=0.47]{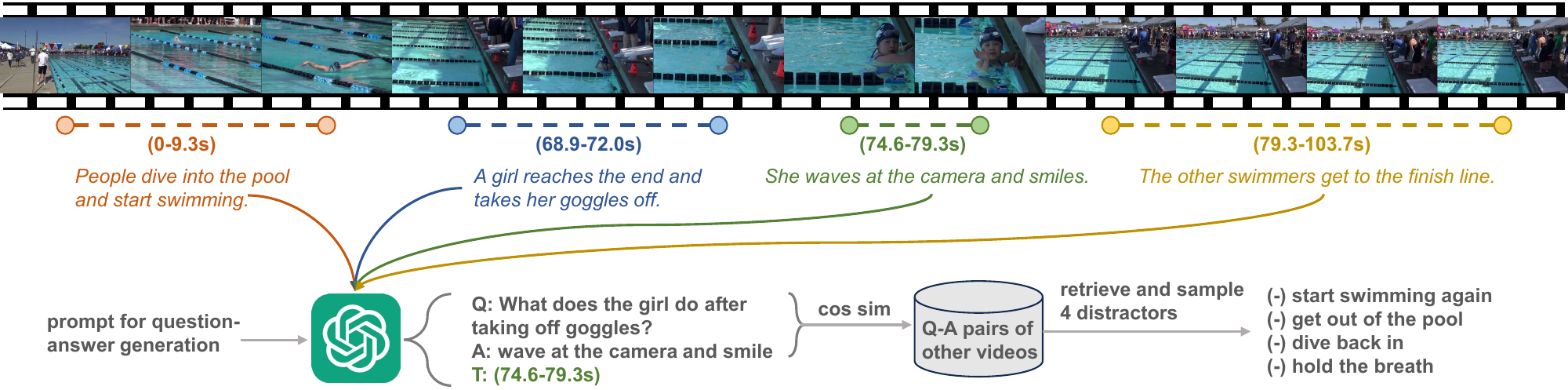}
\end{center}
\caption{Examples of annotation pipeline and generated data for Grounded VideoQA.}
\label{fig:data}
\end{figure*}

\section{Progressive Training}
\label{sec: training}

Different from previous methods \citep{video-llama, video-llava} that train models from scratch using mixed image and video datasets, we start with a pre-trained image-based MLLM \citep{phi3.5-vision} and progressively enhance its fine-grained temporal grounding capabilities. This strategy can be applied to any off-the-shelf MLLM and is more efficient. Table \ref{tab:data} enumerates the datasets used at different stages.

\textbf{Stage-1: Video-Caption Alignment.} We leverage approximately 1.28 million video-text pairs sampled from diverse sources \citep{internvid, webvid, panda70m} to align video encoder’s features with the MLLM. This alignment allows the MLLM, which was pre-trained solely on images, to gain a foundational understanding of videos. Since 2D down-sampling has been applied to the visual features, only the two projectors $f(\cdot)$ and $g(\cdot)$ are trainable, while the video encoder, image encoder, and LLM remain frozen. As this stage does not involve any temporal grounding tasks, the temporal tokens in Sec.\ref{subsec: temporal token} are not yet incorporated.

\textbf{Stage-2: Temporal Token Alignment.} While Video-Caption Alignment connects videos and the MLLM at a coarse level, a gap persists between this alignment and fine-grained temporal grounding. To address this, we introduce the temporal tokens described in Sec.\ref{subsec: temporal token} and continue pre-training the model on a diverse range of grounding datasets \citep{vtimellm, momentor, hawkeye}, focusing on tasks such as Temporal Sentence Grounding, Dense Video Captioning, and Temporal Referring, which enables the model to refer to and localize temporal information effectively. Since new tokens are introduced, we additionally make the word embedding matrix and the final classifier head of the LLM trainable. This stage enhances the model's ability to comprehend multiple events and aligns the temporal tokens with both the video timelines and the LLM's semantic space as shown in Figure \ref{fig:qualitive} and Table \ref{tab:ablation_two_stream}, distinguishing us from previous works \cite{lita, vid2seq, momentor}.

\textbf{Stage-3: Multi-Task Instruction Tuning.}
Following the initial two stages, the model has developed a basic understanding of video content and can locate specific timestamps. In this stage, we further enhance the model's fine-grained temporal grounding while improving its responsiveness to diverse instructions. To achieve this, we gather two types of datasets: (1) We compile a wide range of datasets for temporal grounding tasks, similar to Time-IT \citep{timechat} and VTG-IT \citep{vtgllm}, which include tasks of dense video captioning (remove ActivityNet \citep{activitynet}), temporal sentence grounding, and grounded VideoQA. (2) We incorporate a subset of instructional datasets from VideoChat2 \citep{mvbench} and ShareGPT-4Video \citep{sharegpt4video} to further enhance the model’s ability to generate detailed video captions. By utilizing these diverse datasets, which encompass both temporal grounding and video instruction tasks, \textit{Grounded-VideoLLM} excels in temporal referring, localization, reasoning, and general comprehension of video content. In this stage, the trainable parameters remain the same as in Stage 2, with the addition of LoRA parameters \citep{lora} for the LLM.

\begin{table*}[t!]\scriptsize
  \centering
    \resizebox{\linewidth}{!}{
    \begin{tabular}{l|c|ccc|c|ccc|c|cc}
    \toprule
    \multirow{2}{*}{Model} &\multicolumn{1}{|c}{LLM} &\multicolumn{4}{|c}{Charades-STA} &\multicolumn{4}{|c}{ActivityNet-Grounding} &\multicolumn{2}{|c}{ActivityNet-Captions}\\
    \cmidrule(lr){3-6} \cmidrule(lr){7-10} \cmidrule(lr){11-12}
    &Scale &R@0.3 &R@0.5 &R@0.7 &mIoU &R@0.3 &R@0.5 &R@0.7 &mIoU &SODA\_c &METEOR\\
    \midrule 
    Video-LLaMA \citep{video-llama}  &7B   & 25.2 & 10.6 & 3.4  & \cellcolor{gray!10}{16.8} & 21.9 & 10.8 & 4.9  & \cellcolor{gray!10}{16.5} & 1.9 & 1.9\\
    SeViLA \citep{sevila}            &3B   & 27.0 & 15.0 & 5.8  & \cellcolor{gray!10}{18.3} & 31.6 & 19.0 & 10.1 & \cellcolor{gray!10}{23.0} & -   & -  \\
    Video-ChatGPT \citep{video-chatgpt} &7B  & 27.2 & 6.2  & 1.9  & \cellcolor{gray!10}{19.7} & 19.5 & 10.6 & 4.8  & \cellcolor{gray!10}{14.2} & 1.9 & 2.1\\
    Valley \citep{valley}            &7B    & 28.4 & 1.8  & 0.3  & \cellcolor{gray!10}{21.4} & 30.6 & 13.7 & 8.1  & \cellcolor{gray!10}{21.9} & 0.3 & 0.8\\
    VideoChat \citep{videochat}      &7B   & 32.8 & 8.6  & 0.0  & \cellcolor{gray!10}{25.9} & 23.5 & 12.6 & 6.0  & \cellcolor{gray!10}{17.4} & 0.9 & 0.9\\
    Momenter \citep{momentor}        &7B  & 42.6 & 26.6 & 11.6 & \cellcolor{gray!10}{28.5} & 42.9 & 23.0 & 12.4 & \cellcolor{gray!10}{29.3} & 2.3 & 4.7\\
    VTimeLLM \citep{vtimellm}        &7B  & 51.0 & 27.5 & 11.4 & \cellcolor{gray!10}{31.2} & 44.0 & 27.8 & 14.3 & \cellcolor{gray!10}{30.4} & 5.8 & \textbf{6.8}\\
    TimeChat \citep{timechat}        &7B   & -    & 32.2 & 13.4 & -    & -    & -    & -    & -    & -   & -  \\
    VTG-LLM \citep{vtgllm}           &7B   & -    & 33.8 & 15.7 & -    & -    & -    & -    & -    & -   & -  \\
    HawkEye \citep{hawkeye}          &7B  & 50.6 & 31.4 & 14.5 & \cellcolor{gray!10}{33.7} & \textbf{49.1} & \underline{29.3} & 10.7 & \cellcolor{gray!10}{32.7} & -   & -  \\
    \midrule 

    \textit{Grounded-VideoLLM} (Vicuna)   &7B    &\underline{51.8} &\underline{34.3} &\underline{18.3} &\cellcolor{gray!10}{\underline{34.7}} &43.9 &29.1 &\underline{18.3} &\cellcolor{gray!10}{34.5} &\textbf{6.2} &\underline{6.4} \\

    \textit{Grounded-VideoLLM} (Phi3.5)   &4B    & \textbf{54.2} & \textbf{36.4} & \textbf{19.7} & \cellcolor{gray!10}{\textbf{36.8}} & \underline{46.2} & \textbf{30.3} & \textbf{19.0} & \cellcolor{gray!10}{\textbf{36.1}} & \underline{6.0} & \textbf{6.8} \\ 
    
    \bottomrule 
\end{tabular}
}
\caption{Zero shot results on temporal sentence grounding and dense video captioning tasks.}
\label{tab:results_tsg_dvc}
\end{table*}

\begin{table}[t!]
  \centering
    \resizebox{\linewidth}{!}{
    \begin{tabular}{l|c|cccc}
    \toprule
    Model &Acc@GQA &mIoP &IoP@0.5 &mIoU &IoU@0.5 \\
    \midrule 
    VIOLETv2 \citep{violetv2}         & \cellcolor{gray!10}{12.8} & 23.6 & 23.3 & 3.1  & 1.3 \\  
    Temp[CLIP] NG+ \citep{nextgqa}    & \cellcolor{gray!10}{16.0} & 25.7 & 25.5 & 12.1 & 8.9 \\  
    SeViLA \citep{sevila}             & \cellcolor{gray!10}{16.6} & 29.5 & 22.9 & \textbf{21.7} & 13.8\\
    LangRepo \citep{langrepo}         & \cellcolor{gray!10}{17.1} & 31.3 & 28.7 & 18.5 & 12.2 \\ 
    FrozenBiLM NG+ \citep{frozenbilm} & \cellcolor{gray!10}{17.5} & 24.2 & 23.7 & 9.6  & 6.1 \\  
    VideoStreaming \citep{streaming}  & \cellcolor{gray!10}{17.8} & 32.2 & 31.0 & 19.3 & 13.3 \\  
    LLoVi  \citep{llovi}              & \cellcolor{gray!10}{\underline{24.3}} & \textbf{37.3} & \textbf{36.9} & 20.0 & 15.3\\  
    \midrule 
    \textit{Grounded-VideoLLM} (Vicuna)       & \cellcolor{gray!10}{24.0} &32.2 &31.2 &\underline{20.8} &\underline{16.9}\\  
    \textit{Grounded-VideoLLM} (Phi3.5)       & \cellcolor{gray!10}{\textbf{26.7}} & \underline{34.5} & \underline{34.4} & 21.1 & \textbf{18.0} \\  
    \bottomrule 
\end{tabular}
}
\caption{Results on NExT-GQA. Acc@GQA is defined as the percentage of questions that are both correctly answered and visually grounded with IoP $\geq$ 0.5.}
\label{tab:results_nextgqa}
\end{table}

\section{Grounded VideoQA Dataset Generation} 
Grounded VideoQA requires the model to not only answer questions but identify timestamps that support predicted answers, demonstrating the temporal reasoning abilities. NExT-GQA \citep{nextgqa} was manually developed by extending NExT-QA \citep{nextqa} with temporal labels for start and end timestamps. However, annotating these labels is labor-intensive and time-consuming, limiting NExT-GQA only to QA pairs for the validation and test sets. To create a scalable training dataset, as depicted in Figure \ref{fig:data}, we utilized OpenAI GPT-4 \citep{gpt4} to assist in constructing training sets for the grounded VideoQA task. These sets were built on public datasets that already contain temporal labels, such as QVHighlights \citep{qvhighlights}. We framed the task as a multiple-choice VideoQA using a two-round conversational format as depicted in Appendix \ref{appendix: dataset}.

\section{Experiments}
\textbf{Implementation Details.} 
We select Phi3.5-V-Instruct-3.8B \citep{phi3.5-vision} as the base MLLM of \textit{Grounded-VideoLLM}. We also build another \textit{Grounded-VideoLLM} based on LLaVA-1.5-7B \citep{liu2024llava1.5} using Vicuna-1.5 \citep{vicuna2023} as the LLM for fair comparison. For temporal stream, we adopt InternVideo2-1B \citep{internvideo2} as the video encoder. Each video is sampled as a sequence of $T=96$ frames, which are evenly divided into $K=12$ segments. We set the pooling size $\sigma_{S}=2$ for the spatial stream ($N_S=144$ tokens per frame) while $\sigma_{T}=4$ ($N_T=16$ tokens per frame) for the temporal stream. Moreover, we introduce $M=300$ temporal tokens into the LLM's vocabulary for timestamp representation. We use the Phi3.5 version for ablations in Section \ref{sec:ablation}. More details are in Appendix \ref{appendix: implementation}.

\textbf{Tasks and Benchmarks.} 
 We assess \textit{Grounded-VideoLLM} across three video temporal grounding tasks: \textit{Temporal Sentence Grounding}, \textit{Dense Video Captioning}, and \textit{Grounded VideoQA}, utilizing benchmarks such as \textbf{Charades-STA} \citep{charades-sta}, \textbf{ActivityNet-Captions} \citep{activitynet}, and \textbf{NExT-GQA} \citep{nextgqa}. We also show its reasoning capability by the task of Open-Ended VideoQA with benchmarks including \textbf{MSVD-QA}, \textbf{MSRVTT-QA} \citep{msqa}, and \textbf{ActicityNet-QA} \citep{activityqa}. Additionally, to evaluate the model’s general video understanding capabilities, we benchmark \textit{Grounded-VideoLLM} against existing models using \textbf{VCG-Bench} \citep{video-chatgpt} and \textbf{MVBench} \citep{mvbench}. The evaluation details are in Appendix \ref{appendix: evaluation}.

\subsection{Main Results}

\textbf{Temporal Sentence Grounding.} As shown in Table \ref{tab:results_tsg_dvc}, \textit{Grounded-VideoLLM} (Phi3.5) achieves "mIoU" scores of 36.8 on Charades-STA and 36.1 on ANet-Grounding, respectively. This performance surpasses previous SoTA end-to-end Video-LLMs, such as HawkEye \citep{hawkeye}, by a substantial margin of +3.4. It is worth emphasizing that this promising "mIoU" performance is largely attributed to significant gains in the "R@0.7" metric compared to other thresholds, demonstrating that \textit{Grounded-VideoLLM} is more advanced in localizing specific moments with finer granularity. Interestingly, although \textit{Grounded-VideoLLM} (Vicuna) may be larger than the Phi3.5 version in terms of model parameters, its overall performance is slightly lower. This is because the base Vicuna model is inherently weaker than Phi3.5. Nevertheless, when using the same LLM as the base model, \textit{Grounded-VideoLLM} (Vicuna) still outperforms other models like TimeChat and VTimeLLM.

\textbf{Dense Video Captioning.} We evaluated \textit{Grounded-VideoLLM} on the ANet-Captions, and the results in Table \ref{tab:results_tsg_dvc} show that \textit{Grounded-VideoLLM} (Phi3.5) achieves the highest SODA\_c score of 6.0, which demonstrates that, thanks to the Temporal Token Alignment training stage, \textit{Grounded-VideoLLM} is highly effective in identifying the multi-event structure of the video and capturing complete storylines. The highest METEOR score (6.8) also indicates that \textit{Grounded-VideoLLM} provides more detailed event descriptions compared with other Video-LLMs.

\textbf{NExT-GQA} \citep{nextgqa} requires the model to both correctly answer questions and provide timestamps that support the answers, highlighting the temporal reasoning capability. According to Table \ref{tab:results_nextgqa}, \textit{Grounded-VideoLLM} achieves the highest Acc@GQA (26.7, +2.4) and delivers comparable IoU and IoP scores to models such as SeViLA \citep{sevila} and LLoVi \citep{llovi}, which use specialized grounding modules or rely on proprietary large language models \citep{gpt4}. The highest Acc@GQA score further demonstrates \textit{Grounded-VideoLLM}'s capability in both fine-grained temporal grounding and high-level reasoning.

\textbf{Open-Ended VideoQA.}
As shown in Table \ref{tab:results_qa_vcg_mvbench}, \textit{Grounded-VideoLLM} achieves state-of-the-art or comparative performance across MSVD-QA, MSRVTT-QA \citep{msqa}, and ActivityNet-QA \citep{activityqa}, highlighting its advancements in general video question answering.

\textbf{General Video-LLM Benchmarks.}
While \textit{Grounded-VideoLLM} excels in fine-grained temporal grounding, we should ensure that it maintains performance in general video understanding. Therefore, we conducted a comprehensive evaluation using VCG-Bench \citep{video-chatgpt} and MVBench \citep{mvbench}. As shown in Table \ref{tab:results_qa_vcg_mvbench}, \textit{Grounded-VideoLLM} achieves promising results in VCG-Bench, with an average score of 3.24, outperforming other Video-LLMs \textit{with temporal grounding capabilities} (e.g., LITA, VTimeLLM). Notably, \textit{Grounded-VideoLLM} surpasses all other Video-LLMs on the TU (Temporal Understanding) task (see Appendix Table \ref{tab:results_vcg}), with a score of 3.12 (+7\%), demonstrating its superior temporal understanding, which can be attributed to the two-stream architecture that can capture motion dynamics. For MVBench which provides 4,000 QA pairs spanning a wide range of scenes categorized into 20 fine-grained tasks, the results, presented in Table \ref{tab:results_qa_vcg_mvbench}, show that \textit{Grounded-VideoLLM} achieves an average score of 60.0, surpassing other Video-LLMs.

\begin{table}[t!]
  \centering
    \resizebox{\linewidth}{!}{
    \begin{tabular}{l|cc|cc|cc|c|c}
    \toprule
    \multirow{2}{*}{Model} &\multicolumn{2}{|c}{MSVD-QA} &\multicolumn{2}{|c}{MSRVTT-QA} &\multicolumn{2}{|c}{ANet-QA} &\multicolumn{1}{|c}{VCG-Bench} &\multicolumn{1}{|c}{MVBench}\\
    \cmidrule(lr){2-3} \cmidrule(lr){4-5} \cmidrule(lr){6-7} \cmidrule(lr){8-9}
    &Acc. &Score &Acc. &Score &Acc. &Score &Avg. &Avg.\\
    \midrule 
    \multicolumn{9}{l}{\textcolor{gray}{\textit{Video-LLMs w/o temporal grounding capability.}}}\\
    Video-LLaMA \citep{video-llama}     & 51.6 & 2.5 & 29.6 & 1.8 & 12.4 & 1.1 & 1.98 & 34.1\\  
    Video-ChatGPT \citep{video-chatgpt} & 64.9 & 3.3 & 49.3 & 2.8 & 35.2 & 2.7 & 2.42 & 32.7\\
    Video-LLaVA \citep{video-llava}     & 70.7 & 3.9 & 59.2 & 3.5 & 45.3 & 3.3 & - & 43.0   \\
    Vista-LLaMA \citep{vista-llama}     & 65.3 & 3.6 & 60.5 & 3.3 & 48.3 & 3.3 & 2.57 & - \\
    MovieChat \citep{moviechat}         & 75.2 & 3.8 & 52.7 & 2.6 & 45.7 & \underline{3.4} & 2.67 & - \\
    LongVLM \citep{longvlm}             & 70.0 & 3.8 & 59.8 & 3.3 & 47.6 & 3.3 & 2.89 & - \\
    VideoChat2 \citep{mvbench}          & 70.0 & \underline{3.9} & 54.1 & 3.3 & 49.1 & 3.3 & 2.98 & 51.1\\
    Chat-UniVi \citep{chatunivi}        & 65.0 & 3.6 & 54.6 & 3.1 & 45.8 & 3.2 & 2.99 & - \\
    P-LLaVA-7B \citep{pllava}           & \textbf{76.6} & \textbf{4.1} & \underline{62.0} & \underline{3.5} & \underline{56.3} & \textbf{3.5} & 3.12 & 46.6\\
    ST-LLM \citep{st-llm}               & 74.6 & \underline{3.9} & \textbf{63.2} & 3.4 & 50.9 & 3.3 & 3.15 & 54.9 \\
    VideoGPT+ \citep{videogpt+}          & -    & -   & -    & -   & -    & - & \textbf{3.28} & \underline{58.7}\\
    \multicolumn{9}{l}{\textcolor{gray}{\textit{Video-LLMs w/ temporal grounding capability.}}}\\  
    TimeChat \citep{lita}                   & -    & -   & -    & -   & -    & -  & - & 38.5\\
    Momentor \citep{momentor}           & 68.9 & 3.6 & 55.6 & 3.0 & 40.8 & 3.2 & -    \\
    VTimeLLM \citep{vtimellm}           & -    & -   & -    & -   & -    & -   & 2.85 \\
    LITA \citep{lita}                   & -    & -   & -    & -   & -    & -  & 3.04 \\
    \midrule 
    \textit{Grounded-VideoLLM} (Vicuna)     & 74.7   & \underline{3.9}  & 61.9    & \textbf{3.6}   & 55.7    & \underline{3.4}  & \underline{3.26}  & 58.1 \\ 
    \textit{Grounded-VideoLLM} (Phi3.5)       & \underline{76.3} & \textbf{4.1} & 60.3 & \textbf{3.6} & \textbf{56.8} & \textbf{3.5} & 3.24 & \textbf{60.0}\\ 
    \bottomrule 
\end{tabular}
}
\caption{Results on VideoQA, VCG-Bench and MVBench. Refer to Appendix Table \ref{tab:results_mvbench}, \ref{tab:results_vcg} for details.}
\label{tab:results_qa_vcg_mvbench}
\end{table}

\begin{figure*}[h]
\label{fig:qualitive}
\begin{center}
\includegraphics[scale=0.45]{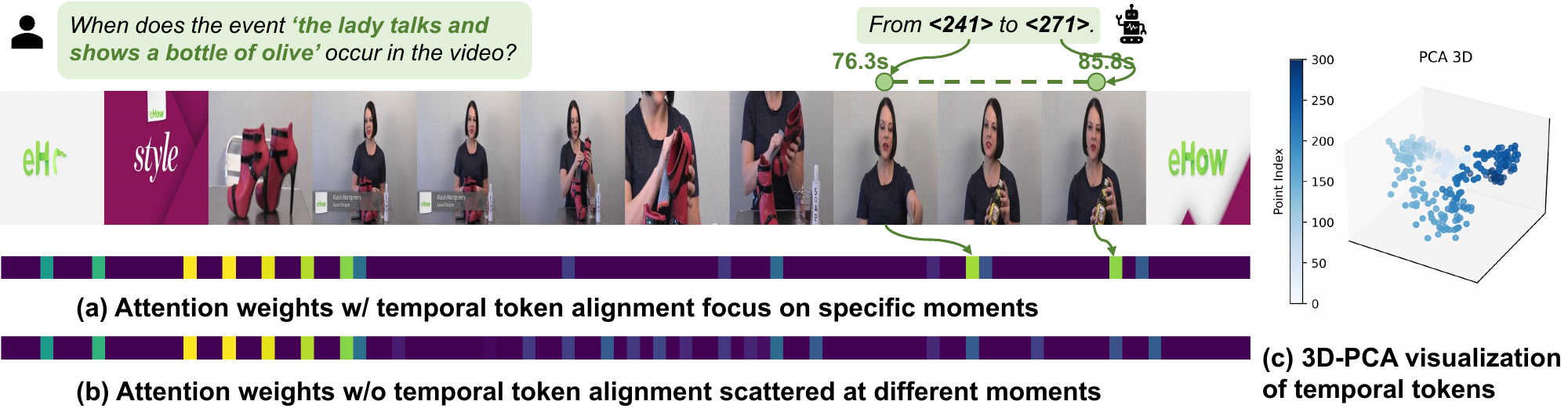}
\end{center}
\caption{Attention weights of the LLM when generating the temporal tokens and 3D-PCA of embeddings.}
\label{fig:qualitive}
\end{figure*}

\subsection{In-Depth Analysis}
\label{sec:ablation}

\textbf{Two-stream Encoding.} We conduct ablations to our two-stream encoding. Specifically, we set three variants by removing the temporal stream or spatial stream respectively: (1) \textit{w/o temporal-stream (dense)} feeds $T=96$ frames with a pooling size $\sigma_{S}=4$ (36 tokens per frame). (2) \textit{w/o temporal-stream (sparse)} feeds $T=24$ frames with a pooling size $\sigma_{S}=2$ (144 tokens per frame). (3) \textit{w/o spatial-stream} feeds $T=14$ frames without pooling (256 tokens per frame). All these variants have a close number of video tokens compared to our two-stream encoding ($12\times(144+\frac{96}{12}\times16)=3264$ tokens for $\textbf{F}_{Vid} \in \mathbb{R}^{K \cdot (N_S + \frac{T}{K} \cdot N_T) \times D}$) for a fair comparison. Table \ref{tab:ablation_two_stream} shows that, interestingly, the dense frame variant performs slightly better in temporal grounding tasks while worse in general benchmarks than the sparse frame variant. This can be attributed to that the videos in MVBench are much shorter and emphasize spatial details more. Our two-stream architecture strikes a balance by capturing dense motion dynamics while maintaining essential appearance details.

\textbf{Temporal Tokens.} We conducted ablations on the temporal tokens to study how they will affect the grounding performance. The models in Table \ref{tab:ablation_temporal_token} were trained using only the first two stages and evaluated on grounding tasks. Specifically, 'w/o temporal tokens' refers to directly using plain text to represent absolute timestamps. The results in Table \ref{tab:ablation_temporal_token} show that while the performance of plain text and 100 temporal tokens is comparable, both are outperformed by 300 tokens. Furthermore, the results reveal a consistent improvement in performance with an increasing number of temporal tokens, especially for longer videos (ANet), highlighting the benefit of finer-grained time representations.

\begin{table}[t!]
    \resizebox{\linewidth}{!}{
    \begin{tabular}{l|c|c|c|c|c}
    \toprule
    \multirow{2}{*}{Model} &\multicolumn{1}{|c}{\# of video} &\multicolumn{1}{|c}{C-STA} &\multicolumn{1}{|c}{ANet-G} &\multicolumn{1}{|c}{ANet-Cap} &\multicolumn{1}{|c}{MVBench}\\
    \cmidrule(lr){3-3} \cmidrule(lr){4-4} \cmidrule(lr){5-6}
    &tokens &mIoU &mIoU &SODA\_c &Avg.\\
    \midrule
    \rowcolor{gray!15} \textit{Grounded-VideoLLM} &3264 & \textbf{36.8} & \textbf{36.1} & \textbf{6.0} & \textbf{60.0} \\
    \midrule
    w/o temporal-stream (sparse) &3456 & 30.4 & 28.0 & 4.9 & 58.5 \\
    w/o temporal-stream (dense)  &3456 & 34.3 & 29.2 & 5.4 & 53.2 \\
    w/o spatial-stream  &3584 & 33.5 & 28.7 & 5.5 & 57.7 \\
    w/o temporal token alignment &3264 & 27.5 & 23.1 & 4.7 & 58.9 \\
    \bottomrule
  \end{tabular}
  \caption{Impact of two-stream and token alignment.}
  \label{tab:ablation_two_stream}
  }
\end{table}

\begin{table}[t!]
    \resizebox{\linewidth}{!}{
    \begin{tabular}{l|c|c|c}
    \toprule
    Setting &{C-STA (mIoU)} &{ANet-G (mIoU)} &{ANet-Cap (SODA\_c)}\\
    \midrule
    w/o temporal tokens    & 32.3 & 29.6 & 5.6\\
    w/ 100 temporal tokens & 32.2 & 29.1 & 5.2\\
    w/ 200 temporal tokens & 32.9 & 30.1 & 5.5\\
    \rowcolor{gray!15} w/ 300 temporal tokens & \textbf{33.8} & \textbf{33.1} & \textbf{5.7} \\
    \bottomrule
  \end{tabular}
  \caption{Impact of temporal tokens.}
  \label{tab:ablation_temporal_token}
  }
  \vspace{-5pt}
\end{table}

\textbf{Alignment Training Stage.} We investigate temporal tokens' role by ablating the 2nd training stage of Temporal Token Alignment. Quantitative results in Table \ref{tab:ablation_two_stream} reveal a performance drop across all tasks, particularly in temporal sentence grounding. Upon analyzing the outputs, we found that the model often produces time intervals spanning nearly the entire video (e.g., {from \ttfamily{<0>} to \ttfamily{<300>}}), neglecting the alignment between specific moments and temporal tokens. Qualitatively, we visualize the attention weights of the LLM to demonstrate how temporal tokens attend to corresponding video moments. Details of generating visualizations are provided in Appendix \ref{appendix: visualization}. As shown in Figure \ref{fig:qualitive} (a), when generating the temporal token, e.g. {\ttfamily{<241>}} or {\ttfamily{<271>}}, the attention weights are higher and more focused on their corresponding video moments. Conversely, in Figure \ref{fig:qualitive} (b), when the model is trained without the alignment stage, the attention weights of temporal tokens become significantly dispersed across irrelevant moments. This illustrates the necessity of our multi-stage training strategy for temporal alignment. We also visualize the embedding distribution of temporal tokens with PCA in Figure \ref{fig:qualitive} (c), revealing that temporal tokens with similar indices tend to cluster together, exhibiting a continuous transition from tokens with smaller indices to larger ones.

\begin{table}[t!]\scriptsize
    \resizebox{\linewidth}{!}{
  \begin{tabular}{l|ccc}
    \toprule
    \multirow{2}{*}{Model} &\multicolumn{3}{|c}{NExT-GQA}\\
    \cmidrule(lr){2-4}
    &Acc@GQA &mIoP &mIoU\\
    \midrule 
    \rowcolor{gray!15} \textit{Grounded-VideoLLM} &26.7 &34.5 &21.1 \\
    \midrule
    w/o grounded VideoQA &18.1 ($\downarrow$ 8.6) &22.2 ($\downarrow$ 12.3) &12.9 ($\downarrow$ 8.2) \\
    \bottomrule
  \end{tabular}
  \caption{Impact of grounded VideoQA dataset.}
  \label{tab:ablation_nextgqa}
  }
  \vspace{-5pt}
\end{table}

\textbf{Grounded VideoQA Dataset.}
We validate the role of our constructed grounded-VideoQA by removing it from the stage-3 data. Since the model without training on our dataset usually generates free-form texts when asked to output the timestamps supporting the answer, we reformulate it as a temporal sentence grounding task, where we combine the predicted answer and question into a single sentence and ask the model to localize its timestamps. Table \ref{tab:ablation_nextgqa} suggests that there is a significant performance decrease with regard to Acc@GQA ($\downarrow$ 8.6), mIoP ($\downarrow$ 12.3), and mIoU ($\downarrow$ 8.2), from which we can conclude that our Grounded VideoQA dataset is essential for the model's temporal reasoning capability.

\section{Conclusion}
We present \textit{Grounded-VideoLLM}, a novel Video-LLM that incorporates a two-stream encoding for temporal modeling, along with the temporal tokens for timestamp representation. We employ a multi-stage training scheme, starting with an image-based MLLM and progressively equipping it with fine-grained temporal grounding capabilities. Additionally, we curate a grounded-VideoQA dataset to further enhance the model's temporal reasoning ability. Extensive experiments demonstrate that \textit{Grounded-VideoLLM} not only excels in video temporal grounding tasks but also performs strongly on general video understanding benchmarks.

\section{Limitations}
While Grounded-VideoLLM demonstrates superiority in handling fine-grained temporal grounding, but it still has some inherent limitations for future works. (1) Timestamp Quantization Error: Although discrete temporal tokens are introduced to represent timestamps and accuracy is improved by increasing the number of tokens, minor quantization errors may still be introduced when converting continuous time into discrete tokens. (2) Computational Resource Requirements: The training and inference processes of the model, especially the parts involving two-stream encoding and large-scale language models, may require more computational resources than single-stream.

\bibliography{custom}

\appendix
\clearpage
\setcounter{page}{1}

\newpage

\section{Appendix}

\begin{table}[h]\scriptsize
\resizebox{\linewidth}{!}{
  \centering
    \begin{tabular}{l|ccccc|c}
    \toprule
    \multirow{2}{*}{Model} &\multicolumn{6}{|c}{VCG-Bench}\\
    \cmidrule(lr){2-6}
    &CI &DO &CU &TU &CO &Avg.\\
    \midrule 
    Video-LLaMA \citep{video-llama}     & 1.96 & 2.18 & 2.16 & 1.82 & 1.79 & \cellcolor{gray!10}{1.98} \\  
    Video-ChatGPT \citep{video-chatgpt} & 2.50 & 2.57 & 2.69 & 2.16 & 2.20 & \cellcolor{gray!10}{2.42} \\
    Vista-LLaMA \citep{vista-llama}     & 2.44 & 2.64 & 3.18 & 2.26 & 2.31 & \cellcolor{gray!10}{2.57} \\
    MovieChat \citep{moviechat}         & 2.76 & 2.93 & 3.01 & 2.24 & 2.42 & \cellcolor{gray!10}{2.67} \\
    LongVLM \citep{longvlm}             & 2.76 & 2.86 & 3.34 & 2.39 & 3.11 & \cellcolor{gray!10}{2.89}\\
    VideoChat2 \citep{mvbench}          & 3.02 & 2.88 & 3.51 & 2.66 & 2.81 & \cellcolor{gray!10}{2.98} \\
    Chat-UniVi \citep{chatunivi}        & 2.89 & 2.91 & 3.46 & 2.89 & 2.81 & \cellcolor{gray!10}{2.99} \\
    P-LLaVA-7B \citep{pllava}           & 3.21 & 2.86 & 3.62 & 2.33 & 2.93 & \cellcolor{gray!10}{3.12} \\
    ST-LLM \citep{st-llm}               & 3.23 & 3.05 & \textbf{3.74} & \underline{2.93} & 2.81 & \cellcolor{gray!10}{3.15} \\
    VideoGPT+ \citep{videogpt+}          & \underline{3.27} & \textbf{3.18} & \textbf{3.74} & 2.83 & \textbf{3.39} & \cellcolor{gray!10}{\textbf{3.28}} \\
    VTimeLLM \citep{vtimellm}           & 2.78 & \underline{3.10} & 3.40 & 2.49 & 2.47 & \cellcolor{gray!10}{2.85} \\
    LITA \citep{lita}                   & 2.94 & 2.98 & 3.43 & 2.68 & \underline{3.19} & \cellcolor{gray!10}{3.04} \\
    \midrule 
    \textit{Grounded-VideoLLM}       & \textbf{3.34} & 2.94 & \underline{3.66} & \textbf{3.12} & 3.14 & \cellcolor{gray!10}{\underline{3.24}} \\ 
    \bottomrule 
\end{tabular}
}
\caption{Results on VCG-Bench. VCG-Bench contains five aspects: Correctness of Information (CI), Detail Orientation (DO), Contextual Understanding (CU), Temporal Understanding (TU), and Consistency (CO).}
\label{tab:results_vcg}
\end{table}

\begin{table*}[h]
  \centering
    \resizebox{\linewidth}{!}{
    \begin{tabular}{l|c|cccccccccccccccccccc}
    \toprule
    Model &Avg. &AS &AP &AA &FA &UA &OE &OI &OS &MD &AL &ST &AC &MC &MA &SC &FP &CO &EN &ER &CI \\
    \midrule 
    Otter-V \citep{otter}              & \cellcolor{gray!10}{26.8} & 23.0 & 23.0 & 27.5 & 27.0 & 29.5 & 53.0 & 28.0 & 33.0 & 24.5 & 23.5 & 27.5 & 26.0 & 28.5 & 18.0 & 38.5 & 22.0 & 22.0 & 23.5 & 19.0 & 19.5 \\
    mPLUG-Owl-V \citep{mplug-owl}      & \cellcolor{gray!10}{29.7} & 22.0 & 28.0 & 34.0 & 29.0 & 29.0 & 40.5 & 27.0 & 31.5 & 27.0 & 23.0 & 29.0 & 31.5 & 27.0 & 40.0 & 44.0 & 24.0 & 31.0 & 26.0 & 20.5 & 29.5 \\
    VideoChatGPT \citep{video-chatgpt} & \cellcolor{gray!10}{32.7} & 23.5 & 26.0 & 62.0 & 22.5 & 26.5 & 54.0 & 28.0 & 40.0 & 23.0 & 20.0 & 31.0 & 30.5 & 25.5 & 39.5 & 48.5 & 29.0 & 33.0 & 29.5 & 26.0 & 35.5 \\
    VideoLLaMA \citep{video-llama}     & \cellcolor{gray!10}{34.1} & 27.5 & 25.5 & 51.0 & 29.0 & 39.0 & 48.0 & 40.5 & 38.0 & 22.5 & 22.5 & 43.0 & 34.0 & 22.5 & 32.5 & 45.5 & 32.5 & 40.0 & 30.0 & 21.0 & 37.0 \\
    VideoChat \citep{videochat}        & \cellcolor{gray!10}{35.5} & 33.5 & 26.5 & 56.0 & 33.5 & 40.5 & 53.0 & 40.5 & 30.0 & 25.5 & 27.0 & 48.5 & 35.0 & 20.5 & 42.5 & 46.0 & 26.5 & 41.0 & 23.5 & 23.5 & 36.0 \\
    TimeChat \citep{timechat}          & \cellcolor{gray!10}{38.5} & 40.5 & 36.0 & 61.0 & 32.5 & 53.0 & 53.5 & 41.5 & 29.0 & 19.5 & 26.5 & 66.5 & 34.0 & 20.0 & 43.5 & 42.0 & 36.5 & 36.0 & 29.0 & 35.0 & 35.0 \\
    Video-LLaVA \citep{video-llava}    & \cellcolor{gray!10}{43.0} & 46.0 & 42.5 & 56.5 & 39.0 & 53.5 & 53.0 & 48.0 & 41.0 & 29.0 & 31.5 & 82.5 & 45.0 & 26.0 & 53.0 & 41.5 & 33.5 & 41.5 & 27.5 & 38.5 & 31.5 \\
    P-LLaVA-7B \citep{pllava}          & \cellcolor{gray!10}{46.6} & 58.0 & 49.0 & 55.5 & 41.0 & 61.0 & 56.0 & 61.0 & 36.0 & 23.5 & 26.0 & 82.0 & 39.5 & 42.0 & 52.0 & 45.0 & 42.0 & 53.5 & 30.5 & 48.0 & 31.0 \\
    VideoChat2 \citep{mvbench}         & \cellcolor{gray!10}{51.1} & 66.0 & 47.5 & 83.5 & 49.5 & 60.0 & 58.0 & 71.5 & 42.5 & 23.0 & 23.0 & 88.5 & 39.0 & 42.0 & 58.5 & 44.0 & 49.0 & 36.5 & 35.0 & 40.5 & 65.5 \\
    ShareGPT4Video \citep{sharegpt4video} & \cellcolor{gray!10}{51.2} & 49.5 & 39.5 & 79.5 & 40.0 & 54.5 & 82.5 & 54.5 & 32.5 & 50.5 & 41.5 & 84.5 & 35.5 & 62.5 & 75.0 & 51.0 & 25.5 & 46.5 & 28.5 & 39.0 & 51.5 \\
    ST-LLM \citep{st-llm}              & \cellcolor{gray!10}{54.9} & 66.0 & 53.5 & 84.0 & 44.0 & 58.5 & 80.5 & 73.5 & 38.5 & 42.5 & 31.0 & 86.5 & 36.5 & 56.5 & 78.5 & 43.0 & 44.5 & 46.5 & 34.5 & 41.5 & 58.5 \\
    VideoGPT+ \citep{videogpt+}        & \cellcolor{gray!10}{58.7} & 69.0 & 60.0 & 83.0 & 48.5 & 66.5 & 85.5 & 75.5 & 36.0 & 44.0 & 34.0 & 89.5 & 39.5 & 71.0 & 90.5 & 45.0 & 53.0 & 50.0 & 29.5 & 44.0 & 60.0 \\
    \midrule 
    \textit{Grounded-VideoLLM} &\cellcolor{gray!10}{60.0} &75.0 &75.5 &83.0 &50.0 &63.0 &88.0 &77.0 &37.0 &41.5 &50.0 &91.5 &45.0 &57.5 &82.0 &49.5 &55.0 &45.5 &32.0 &42.0 &59.0 \\
    \bottomrule 
\end{tabular}
}
\caption{Results on MVBench. MVBench contains 20 aspects: Action Sequence (AS), Action Prediction (AP), Action Antonym (AA), Fine-grained Action (FA), Unexpected Action (UA), Object Existence (OE), Object Interaction (OI), Object Shuffle (OS), Moving Direction (MD), Action Localization (AL), Scene Transition (ST), Action Count (AC), Moving Count (MC), Moving Attribute (MA), State Change (SC), Fine-grained Pose (FP), Character Order (CO), Egocentric Navigation (EN), Episodic Reasoning (ER), Counterfactual Inference (CI), and the average of all 20 metrics (Avg).}
\label{tab:results_mvbench}
\end{table*}

\begin{table*}[h]
  \centering
    \resizebox{\linewidth}{!}{
    \begin{tabular}{l|c|ccc|c|ccc|c|cc}
    \toprule
    \multirow{2}{*}{Model} &\multicolumn{1}{|c}{LLM} &\multicolumn{4}{|c}{Charades-STA} &\multicolumn{4}{|c}{ActivityNet-Grounding} &\multicolumn{2}{|c}{ActivityNet-Captions}\\
    \cmidrule(lr){3-6} \cmidrule(lr){7-10} \cmidrule(lr){11-12}
    &Scale &R@0.3 &R@0.5 &R@0.7 &mIoU &R@0.3 &R@0.5 &R@0.7 &mIoU &SODA\_c &METEOR\\
    \midrule 
    \textit{Grounded-VideoLLM}   &4B &70.2 & 55.9 & 33.2 & 49.4 & 64.9 & 47.8 & 30.4 & 47.2 & 6.6 & 6.5 \\  
    \bottomrule 
\end{tabular}
}
\caption{More results on temporal sentence grounding and dense video captioning tasks. We incorporate a subset of Charades-STA and ActivityNet-Captions into the 3rd training stage and achieve better performance.}
\label{tab:app_res_1}
\end{table*}

\begin{table*}[h]\tiny
  \centering
    \resizebox{\linewidth}{!}{
    \begin{tabular}{l|c|cccccc}
    \toprule
    Model &Acc@GQA &mIoP &IoP@0.3 &IoP@0.5 &mIoU &IoU@0.3 &IoU@0.5 \\
    \midrule 
    \textit{Grounded-VideoLLM}       &29.4 & 37.4 & 48.0 & 36.5 & 27.0 & 41.0 & 25.8 \\  
    \bottomrule 
\end{tabular}
}
\caption{Results on NExT-GQA. We incorporate a subset of Charades-STA and ActivityNet-Captions into the 3rd training stage and achieve better performance.}
\label{tab:app_res_2}
\end{table*}

\begin{figure*}[h]
\label{fig:pca}
\begin{center}
\includegraphics[scale=0.49]{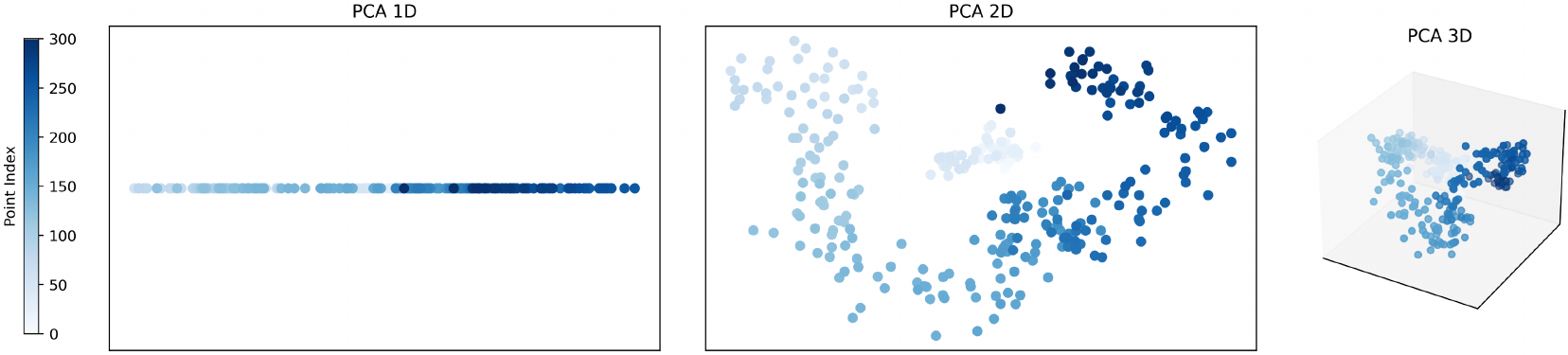}
\end{center}
\caption{Visualization of temporal tokens with PCA.}
\label{fig:pca}
\end{figure*}

\subsection{More Implementation Details}
\label{appendix: implementation}
Phi3.5-Vison-Instruct \citep{phi3.5-vision} consists of a CLIP style ViT image encoder \citep{clip}, an MLP projector $f(\cdot)$, and the large language model Phi3.5-mini-3.8B \citep{phi3}. Each video is sampled as a sequence of $T=96$ frames, which are evenly divided into $K=12$ segments. For the spatial stream encoded by the ViT in Phi3.5-V, we adopt a higher resolution 336$\times$336, but a lower resolution 224$\times$224 for the temporal stream encoded by InternVideo-2. We set the pooling size $\sigma_{S}$ to be 2 while $\sigma_{T}$ to be 4 respectively. For the spatial stream, each frame takes up $24\times24=576$ tokens before while $12\times12=144$ tokens after pooling. For the temporal stream, each frame takes up $16\times16=256$ tokens before while $4\times4=16$ tokens after pooling. Therefore, we have an overall of $K\times(144+\frac{T}{K}\times16)=3264$ tokens in total. During training, we use the AdamW optimizer with a cosine learning rate decay and set the learning rate as 2e-5 and 1e-3 for projector $f(\cdot)$ and $g(\cdot)$ in stage-1. During stage-2 and stage-3, we set the learning rate for both projectors and word embeddings as 2e-5, while 2e-4 for LoRA parameters ($r$ = 128 and $\alpha$ = 256). All experiments are conducted on NVIDIA A100/H800 GPUs.

\subsection{Instructions for Each Task}
The quality and diversity of instructions are essential in the training process. We manually write well-designed instructions for some tasks, combined with some templates in Time-IT \citep{timechat}. Table \ref{tab:task_prompt} lists the prompts for different tasks.

\subsection{Grouned-VideoQA Dataset Generation}
\label{appendix: dataset}
Specifically, we input event descriptions along with their timestamps into GPT-4 and prompted it to generate corresponding question-answer pairs, as shown in Table \ref{tab:gpt4}. To create distractor options for the multiple-choice questions, we retrieved the top 50 questions most similar to the generated question, based on cosine similarity using an embedding model \citep{sentence-embedding}. The answers to these 50 retrieved questions served as candidates for distractors. From this pool, we randomly sampled four distractors with cosine similarities to the correct answer ranging from 0.2 to 0.9, ensuring that the distractors were contextually relevant but not overly similar to the correct answer. The ground-truth timestamps for answering each question were derived from the timestamps of the associated event descriptions. The constructed dataset comprises 17K samples, which have been incorporated into the training sets for Stage 3, further enhancing the model's temporal reasoning performance.

\subsection{Evaluation Process}
\label{appendix: evaluation}
For temporal sentence grounding, we report the metric of Intersection over Union (\textbf{IoU}) \citep{charades-sta} between the timestamps predicted by the model and the ground truth, including \textbf{Recall at IoU} thresholds of \{0.3, 0.5, 0.7\} and their \textbf{mean IoU}. For dense video captioning, we use metrics including \textbf{SODA\_c} \citep{sodac} which is specifically tailored for the video’s storyline, and \textbf{METEOR} score \citep{meteor}, which is the average of traditional METEOR scores that are calculated based on matched pairs between generated events and the ground truth across IoU thresholds of \{0.3, 0.5, 0.7, 0.9\}. For Visually-grounded VideoQA, we calculate both the Intersection of Prediction (\textbf{IoP}) \citep{nextgqa} and Intersection of Union (\textbf{IoU}), and use \textbf{Acc@GQA} \citep{nextgqa} to measure the percentage of questions that are both correctly answered and visually grounded with IoP $\geq$ 0.5. The responses of Open-Ended VideoQA and VCG-Bench are evaluated by GPT-3.5 with the prompts introduced by Video-ChatGPT \citep{video-chatgpt}.

For the evaluation of the temporal sentence grounding task, we directly input the prompt [{\ttfamily{”At which time interval in the video can we see $<query>$ occurring?”}}] in Table \ref{tab:task_prompt} to get the response [{\ttfamily{"From $<start>$ to $<end>$"}}], and then calculate the predicted timestamps with the Equation (\ref{eq:timestamp}) to get the IoU metrics. 

For the evaluation of the dense video captioning task, we directly input the prompt [{\ttfamily{”Detect and report the start and end timestamps of activity events in the video, along with descriptions.”}}] in Table \ref{tab:task_prompt} to get the response [{\ttfamily{"From $<start_1>$ to $<end_1>$, $<caption_1>$. From $<start_2>$ to $<end_2>$, $<caption_2>$. $\cdots$"}}], and then calculate the SODA\_c \citep{sodac} and Meteor scores \citep{meteor}.

For the evaluation of the visually-grounded VideoQA task, we adopt a two-round conversation evaluation: 
\begin{tcolorbox}
\textbf{\textcolor{blue}{Round-1:}}\\
USER: $<question>$. $<options>$. \\
ASSISTANT: Answer: $<answer>$. \\
\textbf{\textcolor{blue}{Round-2:}}\\
USER: Provide the timestamps that correspond to your answer. \\
ASSISTANT: From $<start>$ to $<end>$.
\end{tcolorbox} 
In the first round, we input the question and options into the model and get the answer. In the second round, we input the question, options, and predicted answer as the contexts, together with the prompt [{\ttfamily{"Provide the timestamps that correspond to your answer."}}], into the model to get the predicted timestamps. With both the predicted answers and timestamps, we can calculate the metrics including IoU, IoP, and Acc@GQA \citep{nextgqa}.

For the evaluation of the Open-ended VideoQA and VCG-Bench, we employed GPT-3.5 turbo to juxtapose model outputs with ground truth data as introduced by Video-ChatGPT \citep{video-chatgpt}, subsequently computing both accuracy and a score. To ensure a fair and consistent comparison with the results presented in Video-ChatGPT, we adopted the same prompt for our evaluations. For MVBench, we directly compute the accuracy of multiple-choice questions.

\subsection{Visualization Process}
\label{appendix: visualization}
We visualize the attention weights from the last layer of the LLM during the generation of a new temporal token. Since the full video representation consists of a total of $K \times (N_S + \frac{T}{K} \times N_T)$ tokens—where $T$, $K$, $N_S$, and $N_T$ denote the number of frames, number of segments, number of tokens per frame for the spatial stream, and number of tokens per frame for the temporal stream, respectively—we obtain an attention weight vector with the shape $[K \times (N_S + \frac{T}{K} \times N_T), 1]$. First, we discard the spatial stream portion, focusing only on the temporal information, which results in a new vector with the shape $[K \times \frac{T}{K} \times N_T, 1]$. We then reshape this vector to the form $[T, N_T, 1]$ and average it along the spatial dimension, yielding $[T, 1]$, which represents the final attention weights corresponding to each frame when generating a new token.

\subsection{Distribution of Temporal Tokens}
\label{appendix: pca}
We visualize the embeddings of $M=300$ temporal tokens to investigate their distribution in embedding space. We employ PCA \citep{pca} to reduce the dimensionality of the temporal tokens to 1D, 2D, and 3D representations. For all reductions, we use the reduced values as coordinates, incorporating a gradient color scheme in which the color of the data points changes progressively with the token index, as illustrated in Figure \ref{fig:pca}. Our observations reveal that temporal tokens with similar indices tend to cluster together, exhibiting a continuous transition from tokens with smaller indices (light colors) to those with larger indices (darker colors).

\subsection{More experiment results}
\label{appendix: exp}
As shown in Table \ref{tab:app_res_1}, we incorporate a subset of the training sets of Charades-STA and ActivityNet-Captions into the 3rd training stage and re-train the model from the checkpoint of the 2nd training stage, which achieves better performance. This also greatly improves the performance on NExT-GQA as illustrated in Table \ref{tab:app_res_2}.

\newpage
\begin{table*}
  \centering
  \resizebox{\linewidth}{!}{
    \fcolorbox{black}{gray!10}{
    \parbox{1\linewidth}{
    \textbf{\textcolor{blue}{System}}:
    
      You are a good question generator. I need your help in generating question-answer pairs pertaining to the visual event descriptions. I have a video and I will provide you with descriptions of certain segments and their corresponding timestamps within this video. You need to consider these segments comprehensively based on the given description and timestamps and select one segment which you think can provide a HIGH-QUALITY QUESTION. Based on the description of that segment, ask a question related to that segment, as well as one correct answer. Both the proposed answer and question should be consistent with the content of the give description. BE CAREFUL! Your proposed questions and answers should follow these rules:

    (0) Avoid choosing the segment spanning across the whole video.
    \\
    (1) The question you raised should include causal and temporal relationships as much as possible. Question types should be diverse including WHY, HOW, WHAT, WHERE, etc.
    \\
    (2) NEVER involve anything that is not covered in the given descriptions.
    \\
    (3) The answer should NEVER appear in your question.
    \\
    (4) Your answer should be a phrase no more than 7 words. Keep your answers concise and accurate.
    \\
    
    \textbf{\textcolor{blue}{Demonstrations}}:
    \\
    
    \textbf{User}: 
    \\
    video duration: 82.73 seconds
    \\
    segment-1: [0.83, 19.86] A young woman is seen standing in a room and leads into her dancing.
    \\
    segment-2: [17.37, 60.81] The girl dances around the room while the camera captures her movements.
    \\
    segment-3: [56.26, 79.42] She continues dancing around the room and ends by laying on the floor.
    
    \textbf{Response}: 
    \\
    chosen segment: segment-3
    \\
    segment timestamps: [56.26, 79.42]
    \\
    question: What did the girl do after she ended dancing?
    \\
    answer: lay on the floor
    \\
    
    $\cdots$
    \\
    (other in context demonstrations) 
    \\
    $\cdots$
    }
    }
    }
    \caption{Prompts used to generate visually-grounded VideoQA samples with GPT-4.}
    \label{tab:gpt4}
\end{table*}

\newpage
\begin{table*}
  \centering
  \resizebox{\linewidth}{!}{
    \fcolorbox{black}{gray!10}{
    \parbox{1\linewidth}{
    \textbf{\textcolor{blue}{Prompts for Video-Caption Alignment}}:
    
        (1) "Describe the following video concisely.", \\
        (2) "Provide a brief description of the given video clip.", \\
        (3) "Offer a succinct explanation of the footage presented.", \\
        (4) "Summarize the visual content of the following video.", \\
        (5) "Give a short and clear explanation of the subsequent video clip.", \\
        (6) "Share a concise interpretation of the video provided.", \\
        (7) "Present a compact description of the clip's key features.", \\
        (8) "Relay a brief, clear account of the video shown.", \\
        (9) "Render a clear and concise summary of the video below.", \\
        (10) "Write a terse but informative summary of the following video clip."\\

    \textbf{\textcolor{blue}{Prompts for Temporal Sentence Grounding}}:
    
            (1) "When does $<query>$ happen in the video?",  \\
            (2) "At what time does the occurrence $<query>$ take place in the video?",\\
            (3) "During which part of the video does $<query>$ occur?",\\
            (4) "When in the video does the $<query>$ incident occur?",\\
            (5) "At which moment does $<query>$ take place in the video?",\\
            (6) "During which phase of the video does $<query>$ happen?",\\
            (7) "When does the $<query>$ event occur in the video?",\\
            (8) "At what time does $<query>$ occur in the video sequence?",\\
            (9) "When does the $<query>$ situation take place in the video?",\\
            (10) "At which time interval in the video can we see $<query>$ occurring?"\\

    \textbf{\textcolor{blue}{Prompts for Dense Video Captioning}}:
    
    (1) "Localize a series of activity events in the video, output the start and end timestamp for each event, and describe each event with sentences.",\\
    (2) "Detect and report the start and end timestamps of activity events in the video, along with descriptions.",\\
    (3) "Pinpoint the time intervals of activity events in the video, and provide descriptions for each event.",\\
    (4) "Can you compile a list of the activities and their timestamps featured in the video?",\\
    (5) "I need you to scrutinize the video and catalog every event it contains, along with the timestamps."\\

    \textbf{\textcolor{blue}{Prompts for Temporal Referring}}:
    
    (1) "What is happening from $<start>$ to $<end>$?",\\
    (2) "What is taking place between $<start>$ and $<end>$?",\\
    (3) "What events unfold between $<start>$ and $<end>$?",\\
    (4) "What is happening during the period from $<start>$ to $<end>$?",\\
    (5) "What occurs between $<start>$ and $<end>$?",\\
    (6) "What is going on from $<start>$ to $<end>$?",\\
    (7) "How do things progress from $<start>$ to $<end>$?",\\
    (8) "Can you describe what happens from $<start>$ to $<end>$?",\\
    (9) "Describe the events occurring between $<start>$ and $<end>$.",\\
    (10) "Narrate the actions that unfold from $<start>$ to $<end>$."\\
    }
    }
    }
    \caption{Prompts used for different tasks.}
    \label{tab:task_prompt}
\end{table*}

\end{document}